# AI-RAN: Transforming RAN with AI-driven Computing Infrastructure

Lopamudra Kundu, Xingqin Lin, Rajesh Gadiyar, Jean-Francois Lacasse, and Shuvo Chowdhury

NVIDIA

Emails: {lkundu, xingqinl, rgadiyar, jlacasse, shuvoc}@nvidia.com

*Abstract*— The radio access network (RAN) landscape is undergoing a transformative shift from traditional, communication-centric infrastructures towards converged compute-communication platforms. This article introduces AI-RAN which integrates both RAN and artificial intelligence (AI) workloads on the same infrastructure. By doing so, AI-RAN not only meets the performance demands of future networks but also improves asset utilization. We begin by examining how RANs have evolved beyond mobile broadband towards AI-RAN and articulating manifestations of AI-RAN into three forms: AI-for-RAN, AI-on-RAN, and AI-and-RAN. Next, we identify the key requirements and enablers for the convergence of communication and computing in AI-RAN. We then provide a reference architecture for advancing AI-RAN from concept to practice. To illustrate the practical potential of AI-RAN, we present a proof-of-concept that concurrently processes RAN and AI workloads utilizing NVIDIA Grace-Hopper GH200 servers. Finally, we conclude the article by outlining future work directions to guide further developments of AI-RAN.

## I. INTRODUCTION

Radio access networks (RANs) have been traditionally built as dedicated infrastructures for delivering mobile broadband services to end users [1]. However, as the demands of modern applications and the variety of connected devices have grown rapidly over time, the limitations of conventional RANs have become apparent. As a result, RANs began to evolve beyond mobile broadband, embracing a more versatile design approach to support vertical use cases and heterogeneous workloads [2]. One of the major shifts in RAN's evolution has been towards the convergence of communication and computing, enabling not only connectivity services but also computing services at the network edge, known as edge computing [3].

Edge computing shifts parts of the data/signal processing from the central cloud or user devices to the edge cloud, promising to offer lower-latency response, reduced transport cost, and enhanced privacy protection [4]. A prominent catalyst for advancing edge computing would be to run both RAN functions and artificial intelligence (AI) workloads, especially generative AI (GenAI) based large language models (LLMs), on the same edge infrastructure [5]. This will transform the traditional, single-purpose RAN infrastructure into an overarching AI-RAN architecture that integrates AI and RAN workloads on the same computing platform. The integration can enable zero touch network and service management (ZSM), intelligent radio resource control, and improved spectral efficiency for telecommunication network [6]. Additionally, the integration can support AI-as-a-service (AIaaS) on the same RAN infrastructure, wherein AI workloads can run alongside RAN workloads sharing the same underlying compute resources [7]. Therefore, AI-RAN can not only improve the network's adaptability and efficiency but also allow telecom operators to improve their asset utilization by dynamically allocating resources as needed across RAN and AI workloads.

AI-RAN has gained significant traction in recent years as a transformative solution, with a potential for enhancing RAN performance and improving RAN asset utilization. The vision of an AI-native RAN has motivated researchers to embed AI into the physical (PHY) and medium access control (MAC) layer protocols for 6G air interface [8]. The convergence of communication and computing architecture has become essential for supporting such AI-native RAN functionalities. Literature on communication-compute convergence for RAN, such as [9], has highlighted the importance of close coordination between connectivity services and computing services. At this juncture, industry standardization efforts will play a crucial role in shaping global AI-RAN development and deployment. In [10], the International Telecommunication Union Telecommunication Standardization Sector (ITU-T) has specified the requirements for the coordination of networking and computing in 5G networks and beyond. In [11], the ITU Radiocommunication Sector (ITU-R) has identified "AI and communication" as a key 6G usage scenario, highlighting that 6G will include new capabilities to support AI/compute functionalities, such as computing resource orchestration, distributed AI model training, model sharing and distributed inference. Meanwhile, the 3rd generation partnership project (3GPP) has been incorporating AI functionalities into its latest RAN standards, particularly in the context of 5G-Advanced [12]. The O-RAN Alliance has also been evolving RAN with intelligence by introducing radio intelligent controllers (RICs) and AI-based service management and orchestration [13]. Alongside, they are looking into various use cases as well that can be enabled by O-RAN architecture through efficient compute and communication integration by virtue of O-cloud [14]. To further AI-native RAN, an industry consortium called AI-RAN Alliance was formed in 2024. Leading telecom players are joining hands through AI-RAN Alliance with a shared goal of transforming the next generation accelerated infrastructure.

Albeit the growing momentum of AI-RAN, the existing body of works lacks a unified framework for encompassing the diverse usages of AI in RAN. In this article, we address this gap by articulating manifestations of AI-RAN into various forms in Section II, describing requirements and enablers of AI-RAN in Section III, providing a reference architecture for realizing AI-RAN in Section IV, demonstrating the efficacy of AI-RAN through proof-of-concepts in Section V, and concluding with future scopes for the evolution of AI-RAN in Section VI.

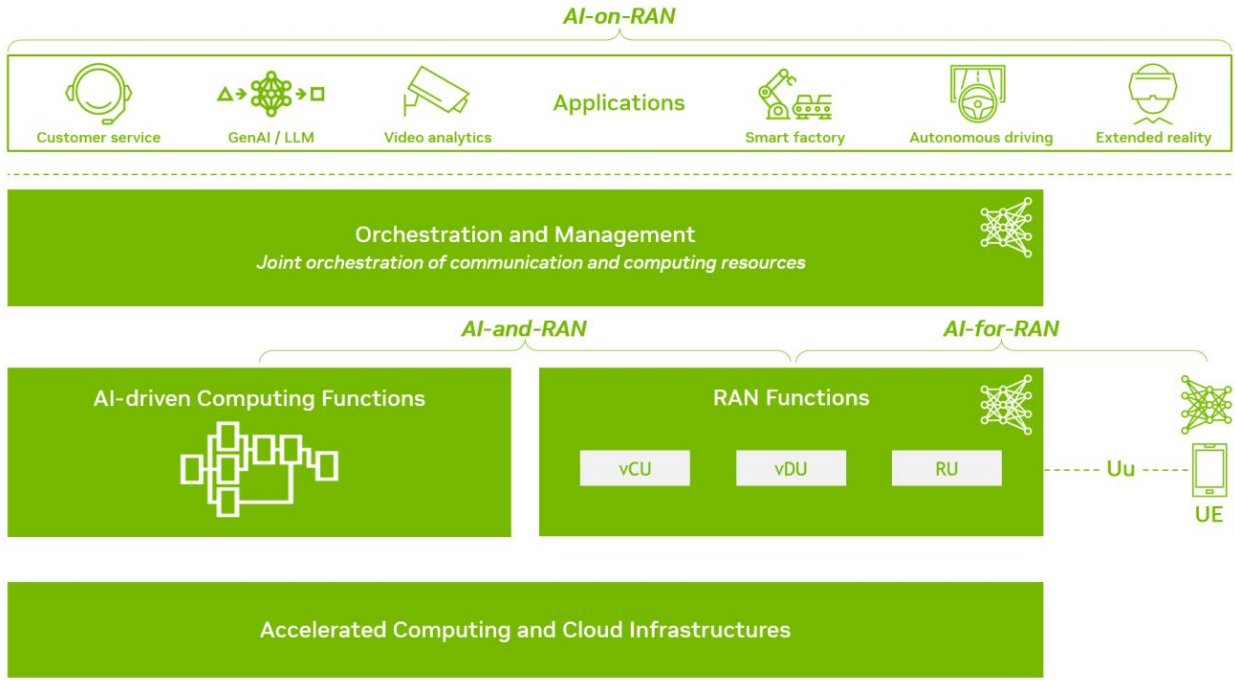

Figure 1: A high-level overview of AI-RAN.

## II. USE CASES OF AI-RAN

Within the unified compute-communication infrastructure of telecommunication network, the integration of AI and RAN workloads can manifest in three tiers – *protocol/architecture level*, *service level* and *infrastructure level*, as illustrated in Figure 1. While global standardization bodies like 3GPP and O-RAN are primarily focusing on protocol/architecture level infusion of AI into RAN, industry initiatives such as AI-RAN Alliance and Global Telecom AI Alliance (GTAA) are underway in exploring other potential avenues of handshaking between AI and RAN to unlock new business opportunities at service and infrastructure levels. Cumulatively, the ongoing activities across various standards development organizations (SDOs) and alliances are marching towards harnessing the power of AI in transforming RAN. Bringing compute and communication together into a common fabric, these initiatives are focusing on three key areas: 1) improving network spectral and operational efficiencies at the protocol/architecture level (*AI-for-RAN*), 2) enabling new AI applications hosted by RAN at the service level (*AI-on-RAN*), and 3) aiding higher asset utilization by sharing compute resources between AI and RAN workloads at the infrastructure level (*AI-and-RAN*) [5].

*AI-for-RAN*: Utilizing AI to enhance RAN performance has been the central theme of AI-native 6G RAN vision [13] and is shaping the standardization trends across multiple SDOs since the era of 5G-advanced [12]. Potentials for AI-driven solutions in improving spectral efficiency, reducing operational cost and elevating user experience have been demonstrated through infusion of AI-based algorithms into the RAN protocol stack – all the way from PHY layer (channel estimation, beamforming optimization, dynamic power adjustment and interference mitigation) to MAC layer and beyond (adaptive modulation and coding (MCS) selection, dynamic resource allocation and load balancing, multi-cell scheduling, and mobility management optimization).

Alongside, network automation has increasingly become crucial with the evolution of virtualized RAN, requiring a high degree of agility, dynamicity and intelligence in network management and orchestration. Managing highly complex RAN across multiple services/applications/users with widely varying performance requirements and usage demands has set forth the journey towards ZSM, for which intent-based management and zero-touch automation are the key enablers. AI-based models, aided by training from historical data and learning from real-time data, are driving this zero-touch transformation through automated user intent interpretation, adaptive decision making, predictive analytics, automated network configuration and service provisioning.

*AI-on-RAN:* Leveraging RAN infrastructure to support AI-based applications is gaining traction in various vertical industry domains such as smart cities and internet of things (IoT), creating a symbiotic relationship between RAN and AI in the telecommunication space. AI-on-RAN can potentially unfold new revenue streams for telecom operators by utilizing the computational assets and capabilities of RAN infrastructure to offer AIaaS to internal operation teams (e.g., customer service, network operations center, marketing, and fraud detection) as well as to external customers (e.g., enterprise).

The notion of AI-on-RAN is rapidly evolving from concept to reality, with leading telecom operators around the globe offering a wide range of GenAI tools, models (e.g., LLMs) and platforms built on their RAN infrastructure [15]. As one example, the founding operator members of GTAA are co-developing a multi-lingual, telco-specific LLM that can be tailored towards personalized customer care services, driving the proliferation of third-party ecosystem. Alongside these standalone GenAI solutions, some telcos are taking a more comprehensive approach of AI-on-RAN, bundling GenAI with other key services like cloud computing and connectivity, such as combined GenAI and 5G solutions offered by Maxis partnered with Amazon Web Services.

*AI-and-RAN*: Enhancing computing resource sharing between RAN network functions (NFs) and AI applications is the crux of sustainable, cost-effective and efficient AI-RAN realization in the telecommunication network. Dynamic orchestration of AI and RAN workloads on the same underlying infrastructure would allow for optimum resource utilization, with computing resources repurposed and shared across multiple workloads as per their computing needs. A practical example of AI-and-RAN solution is Aarna multi cluster orchestration platform (AMCOP) offered by Aarna Networks. Alongside robust automation of RAN workload orchestration on graphics processing unit (GPU) cloud, AMCOP monitors the resource utilization of RAN and dynamically allocates unused GPU cycles to AI applications, improving underlying hardware utilization from an average 30%-40% (running only RAN workload) to nearly 100% (running AI and RAN workloads in parallel). Softbank's recently announced AI-and-RAN solution AITRAS is based on a GPU-based unified infrastructure designed to support multi-tenancy for AI and RAN with AI-based orchestration, enabling telecommunication operators to run AI and RAN workloads concurrently.

No matter at which level and to what extent AI workloads are integrated and co-located with RAN workloads in the manifestation of AI-RAN, it is instrumental to understand the key features of these heterogeneous NFs and applications, so that the underlying platform and its computational capabilities can be optimally architected and provisioned in the RAN infrastructure. In the next section, we will deep dive into these key requirements and enablers of AI-RAN.

## III. REQUIREMENTS AND ENABLERS OF AI-RAN

The design of AI-RAN requires careful consideration of both the computing and the communication functions. The goal is to build a high-performance, scalable, and intelligent RAN platform that combines advanced computational capabilities with optimized communication functions. In this section, we identify key requirements and technology enablers for realizing AI-RAN illustrated in Figure 1.

*Accelerated computing infrastructure*: Traditional RAN computing infrastructure typically utilizes custom-made application-specific integrated circuit (ASIC) for baseband processing and multi-core central processing unit (CPU) for higher layer processing. Such infrastructure cannot handle the complex tasks associated with the diverse AI-RAN use cases. AI-RAN requires general-purpose, software-defined hardware accelerators such as GPUs for handling intensive workloads including not only RAN compute but also AI model training and inference. The infrastructure must be optimized for both training AI models and deploying them for inference at the edge. Offline training of large AI models may happen in the cloud, while online training and inference need to occur in real-time at the edge nodes with integrated computing capabilities.

Furthermore, accelerated computing infrastructure with massively parallel processing is needed to handle the vast amount of data generated in real-time by RAN and AI services. This capability speeds up AI algorithms as well as baseband signal processing, enabling faster computing for workloads such as massive multiple-input multiple output (MIMO) beamforming, multi-cell scheduling, and integrated sensing and communication (ISAC). The infrastructure should also be equipped with high-speed, low-latency interconnects to ensure fast and seamless communication across the network nodes.

*Software-defined, cloud-native design*: Conventional RAN evolution from one generation to another usually triggers a hardware refresh of the RAN infrastructure. Such RAN evolution approach is not sustainable, calling for software-defined, cloud-native design in the emerging AI-RAN paradigm. In software-defined AI-RAN, NFs are abstracted from the underlying hardware, enabling greater flexibility, portability and programmability. Specifically, NFs within the AI-RAN, such as baseband processing, scheduling, and radio resource management (RRM), should be implemented in software, allowing for dynamic instantiation of RAN functions as cloud-native network functions (CNFs). This enables easy updates, reconfigurations, and the ability to introduce new features and services (e.g., emerging edge AI applications) without requiring hardware changes.

AI-RAN should support elastic scaling of both computing and communication resources, adding or removing CNFs based on demand. Cloud-native design is crucial to enable scalability, flexibility, and efficiency. CNFs can be decomposed into microservices, each of which may handle a specific function and adapt to varying workloads. Cloud-native AI-RAN uses containers (e.g., Docker, Kubernetes) to package and deploy CNFs as microservices, enabling efficient use of resources in multi-tenancy (i.e., running multiple RAN and AI services on a shared infrastructure). Continuous integration and continuous delivery or deployment (CI/CD) of RAN and AI services enables rapid deployment of new features and upgrades with minimal service disruption.

*Joint orchestration of communication and computing resources*: The inherent multi-tenant nature of AI-RAN necessitates joint orchestration of communication and computing resources, which calls for a unified orchestration platform. This enables real-time, on-demand allocation of resources based on the current network traffic, user demand, and service-level agreements (SLAs). The orchestrator should manage both computational (e.g., GPU, CPU, memory) and communication (e.g., bandwidth and time) resources, according to the specific requirements of different workloads. It needs to decide dynamically where a specific service or workload should be executed. For example, the orchestrator may assign edge resources to latency-sensitive workloads while routing compute-heavy, non-real-time tasks to centralized cloud nodes. AI models should be integrated into the orchestration platform to make intelligent decisions regarding the optimal distribution of computing and communication resources for time-varying tasks like power optimization and load management.

Elasticity is crucial for handling fluctuations in network conditions and computing demands. The orchestrator should scale resources up or down as needed to handle the varying traffic and computing demands efficiently, both during peak hours and lower-demand periods. It should exploit the underlying hardware's capabilities to achieve dynamic scaling. For example, multi-instance GPU (MIG) technique can be leveraged to segmentize a physical GPU into several hard-partitioned and memory/fault isolated instances, giving the orchestrator the ability to support concurrent processing of RAN workloads and edge AI applications on pooled GPU hardware resources.

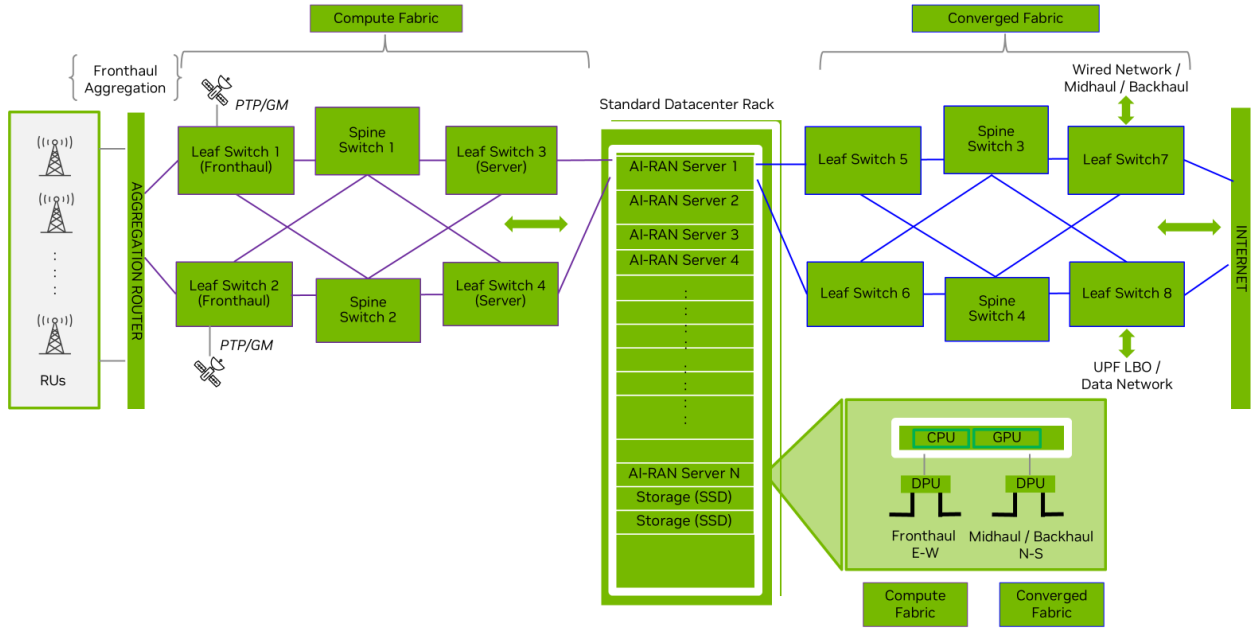

Figure 2: Schematic of AI-RAN reference architecture with server rack and network topology.

*Native AI support*: The emerging AI-RAN paradigm must support AI natively by embedding AI capabilities directly into the RAN infrastructure to enable real-time, intelligent decision-making, automation, and optimization across both the network and the computing layers. From the RAN design perspective, AI should be incorporated into all network layers right from the beginning, spanning all the way from PHY layer processing to higher layer NFs such as RRM, mobility management, load balancing, and scheduling. AI models should be capable of coordinating decisions across different layers of the network to achieve holistic, end-to-end network optimization, where decisions made at one layer take into account the impact on other layers. For accurate, AI-driven optimizations and decisions, the AI-RAN system should have robust data pipelines that collect and process data from various network elements, nodes and layers.

Native AI support in AI-RAN necessitates the use of network digital twin (NDT) which is a virtual replica of the physical network for data collection, simulation, and optimization. The NDT provides a virtual sandbox for training AI models without the risk of disrupting live AI-RAN network operations. Once trained, AI models can be validated in the NDT environment to test their behavior under various conditions. This ensures that AI models perform as expected and do not cause unintended issues like downtime when deployed in the live AI-RAN network.

Bringing all these key features and technologies together will enable the foundation of AI-RAN architecture, based on which telecom operators can build their own, customized AI-RAN deployments augmented on top of their RAN infrastructure. In the next section, we will detail a blueprint of AI-RAN architecture coalescing these fundamental technology enablers.

## IV. REFERENCE ARCHITECTURE DESIGN

The reference architecture of AI-RAN can be built upon the foundational principles of high-performance, scalability, and modularity in AI and RAN computation convergence. To guide AI-RAN deployments, a solution blueprint consisting of standard datacenter rack with AI-RAN servers (with CPUs, GPUs, data processing units (DPUs), and solid-state drives (SSDs)) and ethernet switch based networking fabric is depicted in Figure 2. The schematic provides a reference architecture for telcos to deploy the next-generation, software-defined and accelerated data center for AI-RAN, addressing the computational needs of AI and RAN workloads together. Cloud-native accelerated compute is at the core of this reference architecture, enabling rapid deployment of AI-RAN systems with varying degrees of scaling and computing demands as per the RAN traffic and AI workload coming at the telco edge data centers (such as central offices and mobile switching offices) over time. In the end-to-end AI-RAN deployment blueprint, the key components include radio units (RUs), fronthaul network, distributed units (DUs) (and in some cases, central Units (CUs) and core networks (CNs) as well) running on AI-RAN servers, midhaul, backhaul and AI networks connecting all the way to the internet. Note that, for simplicity, Figure 2 depicts fronthaul network topology connecting RUs to a single AI-RAN server (i.e., many-to-one mapping), whereas in practical deployments the connections between RUs and AI-RAN servers will be many-to-many.

For seamless flow of AI and RAN traffics through the same infrastructure, the networking fabric is divided into two parts, viz., compute fabric (between RUs and AI-RAN servers) and converged fabric (between AI-RAN servers and internet). The compute fabric distributes the RAN workload via fronthaul across AI-RAN servers (east-west (E-W) traffic flow). The converged fabric carries the combined RAN and AI workloads to and from the AI-RAN servers via midhaul/backhaul (north-south (N-S) traffic flow) for wireless traffic, as well as provides connectivity to wired networks for AI traffic that is not originated from wireless network. In Figure 2, both these two fabrics are represented by a minimal two-spine, four-leaf tree topology, which can further scale in real-world deployments. Within the compute fabric, fronthaul connections coming from the RUs are aggregated in a cell site/transport aggregation router and connected to the AI-RAN servers via a 'spine-leaf' networking fabric. This two-switching layer architecture is commonly used in datacenter networking topology for scalability, redundancy, performance, and simplified network management.

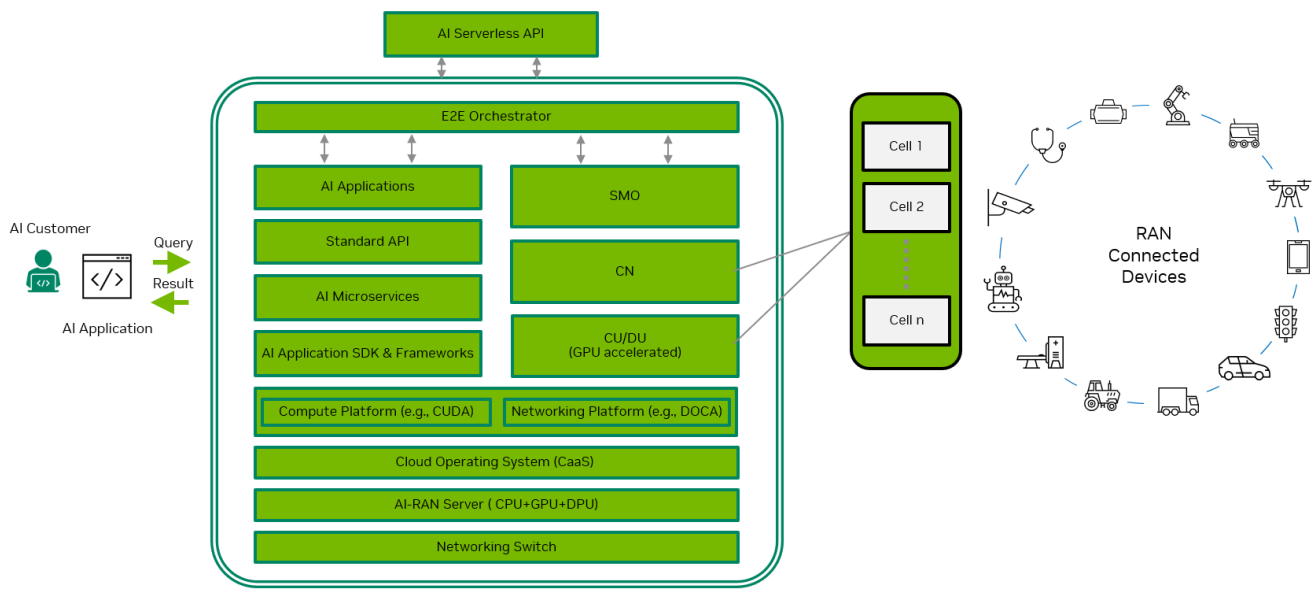

**Figure 3: AI-RAN multi-tenancy software stack.**

In a typical spine-leaf mesh, leaf switches connect directly to network edge endpoints (e.g., servers and other edge devices), aggregating traffic from them before sending to the spine layer, while spine switches form the core of the networking fabric, routing traffic between leaf switches. The compute fabric comprises two types of leaf switches, viz., fronthaul leaf pair switches serving as access points for RUs (e.g., leaf switches 1-2 in Figure 2) and server leaf pair switches (e.g., leaf switches 3-4 in Figure 2) connecting spine layer to edge AI-RAN servers. Each fronthaul leaf switch distributes timing via precision time protocol grandmaster (PTP/GM), the primary source of timing synchronization within the compute fabric of the network. Utilizing PTP protocol, the fronthaul leaf switches distribute precision timing information to RUs connected to the fronthaul network as well as to the DUs in the AI-RAN servers via lower layer split configuration 3 (LLS-C3) synchronization topology as per O-RAN fronthaul specification. The mesh topology created by fronthaul/server leaf switches interconnecting via spine layer creates a highly scalable and redundant network architecture in compute fabric.

While each AI-RAN server connects to the compute fabric at the frontend (i.e., towards the fronthaul network), its backend connects to converged network leaf pairs (e.g., leaf switches 5-8 in Figure 2) interconnected via a mesh of spine switches (e.g., spine switches 3-4 in Figure 2). The converged fabric connects the AI-RAN servers to midhaul, backhaul, or internet depending on whether the AI-RAN servers are hosting only DU, combined DU and CU, or combined DU, CU and CN. For example, AI-RAN servers hosting only DU may connect via midhaul towards CU, whereas AI-RAN servers cohosting DU and CU could connect via backhaul towards user plane function-local breakout (UPF LBO) or towards UPF in CN. A centralized DU+CU+CN running on AI-RAN servers, on the other hand, would connect to the internet (via N6 interface) through the converged fabric, as illustrated in Figure 2.

Next, zooming into the AI-RAN servers in Figure 2, we explore the software stack built upon these servers to support the AI and RAN multi-tenancy on the same platform. Figure 3 illustrates various components of the software stack. It is designed to be cloud-native, with commercial grade cloud operating system (e.g., Kubernetes) offering cloud-as-a-service (CaaS) for dynamic resource orchestration and infrastructure management. Cloud operating system hosts computing platforms and application programming interface (API) models like compute unified device architecture (CUDA) as well as networking platforms and API models like datacenter infrastructure on-a-chip architecture (DOCA) to efficiently run various RAN and AI applications aided by accelerated compute. For the RAN stack, the DU, CU and CN are orchestrated by service management and orchestration (SMO) entity supporting multitudes of cells and RAN applications, whereas for the AI stack, various software components work in concert under API cluster agent that monitors and manages AI server workload in Kubernetes clusters. For the AI stack, the fundamental building blocks comprise AI application software development kit (SDK) and frameworks (e.g., NVIDIA neural modules (NeMo) framework), AI microservices (e.g., NVIDIA inference microservice (NIM)), and industry-standard APIs to connect these components with various AI applications (e.g., text, speech, video, image) running on this platform either natively or through serverless APIs. An overarching end-to-end (E2E) orchestrator simultaneously works with RAN SMO and API cluster agent to track resource utilization and orchestrate RAN workload and AI inferencing requests on the same shared hardware, enabling multi-tenancy while maintaining desired quality-of-service and quality-of-experience requirements for RAN.

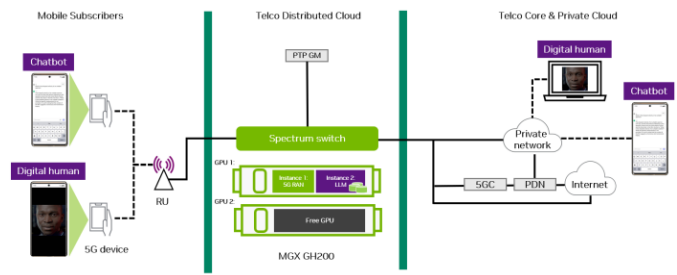

**Figure 4: AI-RAN proof-of-concept schematic diagram.**

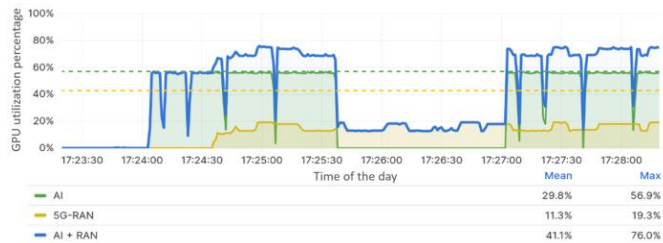

**Figure 5: GPU utilization traces for 1) AI workload, 2) 5G RAN workload, and 3) combined AI and 5G RAN workloads.**

With the AI-RAN reference architecture and the associated software stack, a complete AI-RAN deployment blueprint is available for network operators to enable AI with RAN in the same infrastructure addressing various use case scenarios. One such use case is explained in details using a proof-of-concept demonstration in the following section.

## V. AI-RAN PROOF-OF-CONCEPT

One of the first principles of AI-RAN technology is to be able to run RAN and AI workloads concurrently and without compromising carrier-grade performance. This multi-tenancy can be either in time or in space: dividing the resources based on time of the day, or based on percentage of compute. This also implies the need for an orchestrator that can provision, de-provision, or shift workloads seamlessly based on available capacity as well as dynamic compute need. In this section, we demonstrate the concept of concurrent RAN and AI processing utilizing Grace-Hopper GH200 servers with dynamic allocation of resources between RAN and AI workloads.

In the proof-of-concept experiment depicted in Figure 4, through a 5G network enabled by NVIDIA Aerial, a mobile user interacts with a chatbot or digital human both served by an LLM running on the same server that processes 5G RAN workload. A user can also interact with a chatbot or digital human through telco private network, as shown in the right part of Figure 4. The chatbot implementation consists primarily of a web browser and the LLM, while the digital human implementation further includes voice transcoding, image rendering, and streaming. The LLM, the intelligence driving the responses, is common between the chatbot and digital human implementations, and exists securely at the edge within a telco distributed cloud. The LLM, trained on customer or telco specific data, runs without affecting the RAN workload. The RAN workload involves processing signals associated with a 5G node B (gNB) equipped with 4 transmit antennas and 4 receive antennas (4T4R). The 5G signal has a 100 MHz carrier bandwidth with a 30 kHz subcarrier spacing.

In Figure 4, the MGX GH200 server is equipped with 2 GPUs. MIG technique is used to segmentize the first GPU into two instances: the first GPU instance utilizes up to 40% of the total GPU 1's resources to execute the 5G RAN workload, while the second GPU instance executes AI workloads on the remaining 60% of the total GPU 1's resources. It is noteworthy that in this scenario, because of the use of MIG technique, 5G RAN and AI workloads can be consolidated into GPU 1, freeing up an entire GPU 2 for use by other applications. Figure 5 illustrates the enhanced resource utilization achieved by concurrently running RAN and AI workloads on the same server within the AI-RAN infrastructure. This approach maximizes GPU utilization and minimizes idle states, as opposed to a RAN-only workload that would peak at 40% GPU usage, leaving 60% underutilized. Consequently, this method reduces the total cost of ownership for service providers by enabling the potential to offer unused GPU resources for other services when demand permits.

## VI. CONCLUSION AND FUTURE WORK

In this article, we examined the transformative shift in RANs as they evolve from traditional, communication-centric infrastructures towards converged compute-communication platforms supporting AI-native functionalities. By running both RAN and AI workloads on the same infrastructure, AI-RAN opens the door to more adaptive, efficient, and intelligent networks that dynamically allocate resources to meet the demands of users, services, and emerging vertical industries.

AI-RAN architectures and frameworks are still in their early developmental stages. We conclude here by pointing out some fruitful avenues for future work.

*Advanced orchestration and control*: Future research should focus on designing AI-driven, closed-loop orchestration frameworks that seamlessly manage compute and communication resources. These frameworks must account for latency, reliability, and energy constraints, while providing on-demand resource allocation for both RAN and AI workloads.

*Standardization and interoperability*: Developing and refining open, interoperable interfaces and data formats will be beneficial for the widespread adoption of AI-RAN. Collaborative efforts between standards bodies and industry consortia are needed to ensure that AI-RAN can integrate multi-vendor solutions without compromising performance.

*Testbeds, benchmarks, and real-world trials*: To advance AI-RAN from concept to practice, comprehensive testbeds and benchmarking methodologies are essential. Large-scale simulations, real-time experiments, and pilot deployments will enable stakeholders to validate proposed solutions, understand trade-offs, and refine best practices.

## BIOGRAPHIES

**Lopamudra Kundu** is a Senior Standards Engineer at NVIDIA, leading O-RAN Alliance delegation and driving open RAN ecosystem engagement.

**Xingqin Lin** is a Senior 3GPP Standards Engineer at NVIDIA, leading 3GPP standardization and conducting research at the intersection of 5G/6G and AI.

**Rajesh Gadiyar** is the VP Engineering for Telecom and Edge at NVIDIA, driving innovations at the intersection of 5G/6G, AI and Cloud technologies.

**Jean-Francois Lacasse** is a Developer Relations Manager at NVIDIA, evangelizing and driving the integration of AI and 5G/6G technologies with customers and the broader telco ecosystem.

**Shuvo Chowdhury** is a Principal Product Manager at NVIDIA, responsible for building AI-RAN & Edge AI platforms that focus on the intersection of AI, 5G/6G and Cloud at NVIDIA.